\documentclass[letterpaper, 10 pt, conference]{ieeeconf}  % Comment this line out if you need a4paper

\IEEEoverridecommandlockouts                              % This command is only needed if 
\usepackage{amsmath}
\usepackage{amssymb}
\usepackage{wrapfig}
\usepackage{subfig}
\usepackage{caption}
\usepackage{color}

\usepackage{algorithm}
\usepackage{algorithmic}
\usepackage{graphicx}

\usepackage{pgfplots}
\usepackage{tikz}
\usepackage{tikz-3dplot}

\usepackage{ragged2e}
\usepackage{fp}

\usepackage{ifthen}

\usepackage{tikz}
\usetikzlibrary{shapes.geometric, arrows, calc,automata,positioning,decorations,decorations.text}
\usepackage{multirow}

\usepackage{algorithm}
\usepackage{algorithmic}
\usepackage{graphicx}
\usepackage[english]{babel}
\usepackage{array}
\usepackage[export]{adjustbox}
\usepackage{rotating}

\graphicspath{{images/}}

\usepackage{xcolor}

\usepackage{colortbl}
\usepackage{makecell}
\usepackage{pifont}

\newcommand{\cmark}{\ding{51}}%
\newcommand{\xmark}{\ding{55}}

\title{\LARGE \bf
Semi Supervised Deep Quick Instance Detection and Segmentation}

\author{Ashish~Kumar$^{1}$, L. Behera$^{1}$, \textit{Senior Member IEEE} \\
{\tt\small {\{\color{red!90!blue}https://github.com/ashishkumar822}\}}
\thanks{$^{1}$Mr. Ashish Kumar and Dr. L. Behera are with the Department of Electrical Engineering, Indian Institute of Technology, Kanpur
        {\tt\small \{krashish,lbehera\}@iitk.ac.in}}%
}

\begin{document}

\maketitle
\thispagestyle{empty}
\pagestyle{empty}

\definecolor{skyblue}{rgb}{.529,.807,.980}
\definecolor{airforceblue}{rgb}{0.36, 0.54, 0.66}

\definecolor{bleudefrance}{rgb}{0.19, 0.55, 0.91}
\definecolor{applegreen}{rgb}{0.55, 0.71, 0.0}

\tikzstyle{block} = [rectangle, minimum width=1ex, minimum height=1ex, text centered]

\tikzstyle{circles} = [draw=black, circle, text centered]
\tikzstyle{ellipses} = [draw=black, ellipse, text centered]

\tikzstyle{arrow} = [thin, ->, >=latex]
\tikzstyle{line} = [thin]

\begin{justify}

\begin{abstract}
In this paper, we present a semi supervised deep quick learning framework for instance detection and pixel-wise semantic segmentation of images in a dense clutter of items. The framework can quickly and incrementally learn novel items in an online manner by real-time data acquisition and generating corresponding ground truths on its own. To learn various combinations of items, it can synthesize cluttered scenes, in real time. The overall approach is based on the \texttt{tutor-child} analogy in which a deep network (tutor) is pretrained for class-agnostic object detection which generates labeled data for another deep network (child). The child utilizes a customized convolutional neural network head for the purpose of quick learning. There are broadly four key components of the proposed framework: semi supervised labeling, occlusion aware clutter synthesis, a customized convolutional neural network head, and instance detection. The initial version of this framework was implemented during our participation in Amazon Robotics Challenge (ARC), $2017$. Our system was ranked $3^{rd}$rd, $4^{th}$ and $5^{th}$ worldwide in pick, stow-pick and stow task respectively. The proposed framework is an improved version over ARC'$17$ where novel features such as instance detection and online learning has been added.

\end{abstract}

\end{justify}

%%%%%%%%%%%%%%%%%%%%%%%%%%%%%%%%%%%%%%%%%%%%%%%%%%%%%%%%%%%%%%%%%%%%%%%%%%%%%%%%%
\section{Introduction}
Object detection and semantic segmentation are one of the primary tasks in a visual perception system whether biological or artificial. These tasks have been widely explored in the history of machine vision, however, a drastic performance boost has occurred only in a past few years. Convolutional neural networks (CNNs) \cite{resnet}, \cite{vgg} and advances in the massive parallel computing architectures (GPUs, TPUs) are the primary reasons behind the improvement. However, the CNNs, works well only for the data modality on which they are trained and require large amounts of labeled data as well as training time to achieve that. One of the practical applications of CNN based visual recognition is in warehouse automation process, where a large number of novel items arrives in the warehouses on the daily basis. It becomes challenging to generate labeled images and to retrain the CNN based methods, every time a novel item arrives. Hence, in this paper, we develop an online learning system which can generate labeled data from live camera feeds and incrementally learn the items by clutter synthesis, with no manual quality assurance.  

%Thus, it becomes necessary to address the important questions such as: \textit{i}) Is it possible to train the CNNs in short durations e.g. an hour?, \textit{ii}) If yes, what are the possible ways? and \textit{iii}) Can an online learning system be developed to incrementally learn items? In this paper, we explore possible solutions to the above questions in the context of warehouse automation.
\par
In a broad way, two tasks are manually performed in the warehouses: \textit{i}) picking target items from a storage system called ``rack'' and preparing them for packing, known as \texttt{Pick Task}, and \textit{ii}) picking items from a container called ``tote'' and organizing all of them into a rack, known as a \texttt{Stow Task}. To automate these tasks, Amazon Robotics has previously held Amazon Picking Challenge (APC) in $2015$ and $2016$, and ARC'$17$. Each of the above challenge involved object recognition of around $40$ Amazon provided items (\texttt{known-set}), some having multiple instances. Specifically in ARC'$17$, an additional set of novel items was provided merely $45$ minutes prior to the competition. The challenge involved object recognition in a dense clutter of known and novel items.
%

%
%\begin{figure}[t]
%\centering
%\includegraphics[scale=0.125]{full_system_2.png}
%\caption{Our robotic system during ARC'$17$ performing final task. Image courtesy: Amazon Robotics.}
%\label{fig_full_system}
%\end{figure}
%
\begin{figure}[t]
\centering

\begin{tikzpicture}

\node (full_system) []{\includegraphics[width=35ex,height=26ex]{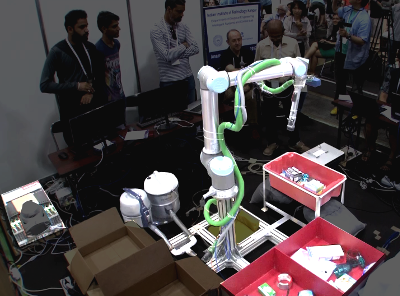}};

%\node (server) [fill=white!100!red, rounded corners=0.3mm, rectangle,text width=12ex,align=center, xshift =-10.5ex, yshift=6ex]{\scriptsize Online learning \\ [-1.3ex]mutli-GPU server};

\node (server) [fill=white!100!red, rounded corners=0.3mm, rectangle,align=center, xshift =-10.5ex, yshift=3ex]{\tiny Mutli-GPU server};

\node (server_dot) [fill=white!0!green, circle, xshift =-14.0ex,yshift=-4.2ex,scale=0.4]{};

\node (server_dot_rect) [draw=white, circle, xshift =-14.0ex,yshift=-4.2ex,scale=0.6]{};

%\node (sunction) [fill=white!100!red, rounded corners=0.3mm, rectangle,align=center, xshift =-6ex,yshift=0ex]{\scriptsize Suction system};
%
%\node (suction_dot) [fill=white!0!green, circle, xshift =-5ex, yshift=-5ex,scale=0.4]{};
%
%\node (suction_dot_rect) [draw=white, circle, xshift =-5ex, yshift=-5ex,scale=0.6]{};

%
%\node (ur10) [fill=white!100!red, rounded corners=0.3mm, rectangle,align=center, xshift =1ex,yshift=10.5ex]{\scriptsize UR10 Arm};

\node (ur10) [fill=white!100!red, rounded corners=0.3mm, rectangle,align=center, xshift =-5ex,yshift=10.5ex]{\tiny UR10 Arm};

\node (ur10_dot) [fill=white!0!green, circle, xshift =0.8ex, yshift=2.5ex,scale=0.4]{};

\node (ur10_dot_rect) [draw=white, circle, xshift =0.8ex, yshift=2.5ex,scale=0.6]{};

\node (ensenso) [fill=white!100!red, rounded corners, rectangle,align=center, xshift =12ex,yshift=10.5ex]{\tiny RGB-D sensor};

\node (ensenso_dot) [fill=white!0!green, circle, xshift =10.3ex, yshift=6.0ex,scale=0.4]{};

\node (ensenso_dot_rect) [draw=white, circle, xshift =10.3ex, yshift=6.0ex,scale=0.6]{};

%\node (ee) [fill=white!100!red, rounded corners, rectangle,align=center, xshift =14ex,yshift=3ex]{\scriptsize End \\[-1 ex] \scriptsize effector};
%
%\node (ee_dot) [fill=white!0!green, circle, xshift = 8.2ex, yshift=3ex,scale=0.4]{};
%
%\node (ee_dot_rect) [draw=white, circle, xshift =8.2ex, yshift=3ex,scale=0.6]{};

\node (tote) [fill=white!100!red, rounded corners, rectangle,align=center, xshift =15ex,yshift=-1.5ex]{\tiny Tote};

\node (tote_dot) [fill=white!0!green, circle, xshift =11ex, yshift=-2.5ex,scale=0.4]{};

\node (tote_dot_rect) [draw=white, circle, xshift =11ex, yshift=-2.5ex,scale=0.6]{};

\node (rack) [fill=white!100!red, rounded corners, rectangle,align=center, xshift =15ex,yshift=-6ex]{\tiny Rack};

\node (rack_dot) [fill=white!0!green, circle, xshift =10.5ex, yshift=-7.5ex,scale=0.4]{};

\node (rack_dot_rect) [draw=white, circle, xshift =10.5ex, yshift=-7.5ex,scale=0.6]{};

%\node (base) [fill=white!100!red, rounded corners, rectangle,align=center, xshift =-10ex,yshift=-11ex]{\tiny Robot base};
%
%\node (base_dot) [fill=white!0!green, circle, xshift =3ex, yshift=-9ex,scale=0.4]{};
%
%\node (base_dot_rect) [draw=white, circle, xshift =3ex, yshift=-9ex,scale=0.6]{};
%

\draw [line width=0.1ex,white] (server) -- (server_dot);
%\draw [line width=0.1ex,white] (sunction) -- (suction_dot);
\draw [line width=0.1ex,white] (ur10) -- (ur10_dot);
\draw [line width=0.1ex,white] (ensenso) -- (ensenso_dot);
%\draw [line width=0.1ex,white] (ee) -- (ee_dot);
\draw [line width=0.1ex,white] (tote) -- (tote_dot);
\draw [line width=0.1ex,white] (rack) -- (rack_dot);
%\draw [line width=0.1ex,white] (base) -- (base_dot);

\FPeval{\segwidth}{18}
\FPeval{\segheight}{12}

\node (img) [xshift=27 ex, yshift=7 ex]{\includegraphics[width=\segwidth ex, height=\segheight ex]{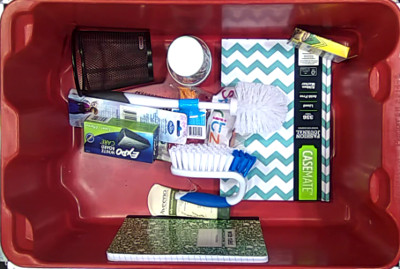}};

\node (img) [xshift=27 ex, yshift=-7 ex]{\includegraphics[width=\segwidth ex, height=\segheight ex]{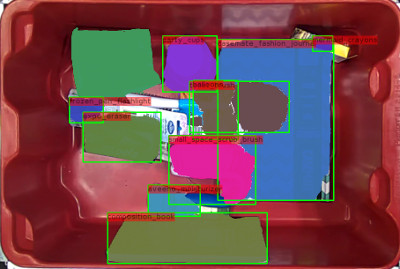}};

\end{tikzpicture}

\caption{Our robotic system during ARC'$17$ performing final task. Image courtesy: Amazon Robotics and Instance detection and segmentation results of $\sim40$ minutes training.}
\label{fig_full_system}
\end{figure}
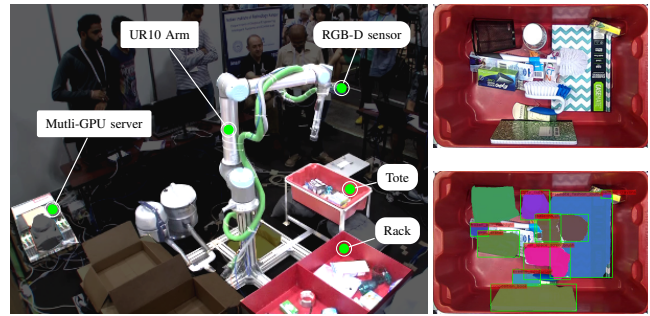

\par
Typically, two approaches can be employed for the pick and stow operations: \textit{i}) pick an item and perform recognition in an isolated view, or \textit{ii}) perform the recognition in the storage system itself and then pick the item.
%Typically, two approaches can be employed for the pick and stow operations: \textit{i}) first pick and then recognize, and \textit{ii}) recognize first and then pick.
 In pick task, the target items may be occluded non-target items. In this case, the first approach becomes suboptimal as the grasping effort and significant amount of time is wasted when a non-target item is picked. On the other hand, the second approach performs equally well for both the stow and pick task. Hence, during ARC'$17$, we opted for in-storage recognition and chose to perform training of CNNs for the desired task in $45$ minutes. This approach, however, demanded acquiring and labeling the images of the novel items in the due time. In addition, the presence of clutter and transparent and wired mesh items made it quite difficult to directly deploy the state-of-art (SOA) methods for the desired purpose.
\par
In this paper, we present a semi supervised quick online learning framework which faces the above challenges without any human intervention. Its inspiration is mainly based on the remarkable capabilities of human to distinguish between seen and unseen classes and differentiating between instances. To enable our system with such a capability, we make use of \texttt{tutor-child} analogy and develop an online learning system where an artificially intelligent (AI) system teaches another AI. The system is capable of learning new items by generating massive synthetic scenes and their ground truths on its own, with real time data acquisition.
\par
 The proposed framework has served as the vision system for our team entry IITK-TCS in ARC'$17$, with a capability of semantic segmentation only. Learning lessons from ARC'$17$, the system in this paper, has been improved to learn while the images are being acquired (online learning) and to detect and segment instances as well. The new functionalities has been added in order to overcome the limitation of semantic segmentation to deal with multiple instances and to reduce manual monitoring. The system components have been designed with an industrial spirit and do not limit their scope only to ARC'$17$. Moreover, due to a large recent attention to the warehouse automation, the underlying ideas of the system components may be concurrently found in the literature, however, their development in this paper is entirely novel.
% The overall algorithm and its ingredients are discussed in Sec.\ref{algorithm}. Their connectivity as an online learning system has been discussed in Sec.\ref{implementation} and the experimental evaluation has been reported in Sec.\ref{experiments}. Next we discuss the related work in area of proposed work and also brief the vision systems developed by other teams.
%

%
\section{Related Work}
Literature on machine vision algorithms is diverse and vast. Therefore, we limit our discussion to object detection and segmentation, and the perception systems developed by other teams. %Therefore, we focus on the SOA methods in the area of object detection and semantic segmentation and brief the perception systems developed by other teams in the context of Amazon challenges.
%\par
Faster-RCNN \cite{fasterrcnn}, Fast-RCNN \cite{fastrcnn}, RCNN \cite{rcnn} are prevalent CNN based approaches to predict rectangular boxes for object detection in the images. RCNN \cite{rcnn} generates object proposal using selective search whereas these are learned as convolutional filters in \cite{fastrcnn}, \cite{fasterrcnn}. FCN \cite{fcn} is the first CNN based approach of end-to-end learning for semantic segmentation. PSPNet \cite{pspnet}, winner of MIT Scene Parsing Challenge $2016$ \cite{mit16scene}, builds upon pyramidal context extraction using average pooling. Mask-RCNN \cite{maskrcnn} integrates both box detection and segmentation techniques for the purpose of instance recognition and segmentation. RefineNet \cite{refinenet} focuses on improving segmentation by aggregating multi-scale information and introducing a boundary refining module.
\par
 APC'$15$ winner, Team RBO \cite{apc15_winner} designed color and depth based features and trained random forest classifiers for pixel-wise segmentation. APC'$16$ winner, Team TuDelft \cite{tudelft2016} employed Faster-RCNN \cite{fasterrcnn}. Team MIT-Princeton \cite{mit2016} in APC'$16$ captured multi-view images to obtain a dense point cloud and used FCN \cite{fcn} for image segmentation. Team Nimbro \cite{nimbro2016} combined object detection and semantic segmentation using RGB-D sensory information. Our Team IITK-TCS in APC'$16$ used Faster-RCNN which led us $5^{th}$ in the stow task. 
\par
ARC'$17$ stow task winner, Team MIT-Princeton \cite{mit2017} first picks an item based on a class agnostic heat map generated by FCN \cite{fcn} and later matches its image in an isolated view with the database using $\ell_{2}$ feature embedding \cite{trippletloss}. The team used $16$ RGB-D sensors which was a quite expensive solution. On the other hand, Team Nimbro \cite{nimbro2017} and the final task winner ACRV \cite{acrvvision2017}, \cite{acrvsystem2017}, similar to us went for in-storage detection. Team Nimbro first obtains masks for novel items using background subtraction and later, trains a ResNeXt-$101$ based RefineNet. Team ACRV also used RefineNet for semantic segmentation. Also, both the teams manually performed quality assurance for automatically generated ground truths by their techniques during the data collection of novel items. In addition, many teams including the above, had constrained their workspaces, in order to cop with the ambient lighting which was a major concern.% On cantrary, our system was an unconstrained workspace and monitor free online learning solution.

\section{Learning Framework}
\label{algorithm}
\label{sec_learning_framework}
Typically, in a \texttt{tutor-child}\footnote{not to be confused with teacher-student networks or Generative Adversarial Networks \cite{gan}} analogy, the tutor gathers data from the world and converts it into a form which can be understood by the child. In our approximation to the above, an AI system (tutor CNN) teaches another AI system (child CNN). To realize this, we develop a semi supervised labeled data generation technique in which single instance images (Fig. \ref{fig_dataset}) are fed to a CNN called tutor which can predict a pixel-wise class agnostic mask of the item present in the image. The predicted mask is converted into a class specific masks by replacing the object mask pixels with the actual label of the item (e.g. brush or bottle), provided by a human. The labeled images are then consumed by a proposed scene synthesis technique in order to synthesize occlusion aware cluttered images. All of the above techniques altogether can be thought as the data conversion process of the tutor-child analogy. Furthermore, the synthesized scenes are fed to the child CNN which employs a novel CNN head to learn quickly for the task of semantic segmentation. The overall flow of the algorithm is shown in Fig. \ref{fig_state_transition} and discussed in detail below.

\subsection{\textbf{Semi Supervised Labeling}}
\label{subsec_auto_anno}
\begin{figure}[t]
\vspace*{0.3ex}

\centering
\subfloat[]
{
\begin{tikzpicture}

\node (ex1) [scale= 0.66]{

\begin{tikzpicture}
\FPeval{\width}{6.7}
\FPeval{\height}{6.7}

\foreach \j in {1,...,3}
\foreach \i in {1,...,3}
\node (node_\i\j) [xshift= (\i-1) * 7 ex + 2 ex,yshift=-(\j-1)*7 ex]{\FPeval{\k}{clip((\j-1)*4 + \i)}
\includegraphics[width=\width ex,height=\height ex]{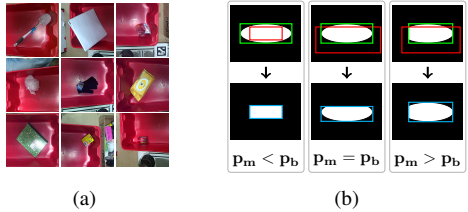}};

%\foreach \j in {4}
%\foreach \i in {1,...,10}
%\node (node_\i) [rectangle, xshift=\i * \width,yshift=-3 * \height]{\FPeval{\k}{clip((\j-2)*10 + \i)}
%\includegraphics[width=\width,height=\height]{images/\k.png}};
\end{tikzpicture}
};
\end{tikzpicture}

\label{fig_dataset}
}
\hspace*{1ex}
\subfloat[]
{
\begin{tikzpicture}

\node (ex1) [scale= 0.64]{
\begin{tikzpicture}
\FPeval{\nodescale}{0.85}

\node (node_less1)[scale=\nodescale]{
\begin{tikzpicture}
\node (rect)[block,fill=black,sharp corners, minimum width=\linewidth, minimum height=0.8*\linewidth,scale=0.2]{};

\node (mask_anno) [ellipses, fill=white, minimum width=8ex, minimum height=3ex, rotate=0]{};
\node (rect_box_anno)[block,thick,draw=red,opacity=1.0, sharp corners, minimum width=5ex, minimum height=2ex]{};

\node (rect_bbox_anno)[block,thick,draw=green,opacity=1.0, sharp corners, minimum width=8ex, minimum height=3ex]{};
\end{tikzpicture}
};

\node (node_less2)[below of=node_less1, yshift=-4ex,scale=\nodescale]{
\begin{tikzpicture}

\node (rect_1)[block,fill=black,sharp corners, minimum width=\linewidth, minimum height=0.8*\linewidth,scale=0.2]{};
\node (rect_box_anno)[block, thick,draw=cyan,fill=white,opacity=1.0, sharp corners, minimum width=5ex, minimum height=2ex]{};

\end{tikzpicture}
};

\node (node_equal1)[scale=\nodescale, xshift=12.5ex]{
\begin{tikzpicture}
\node (rect)[block,fill=black,sharp corners, minimum width=\linewidth, minimum height=0.8*\linewidth,scale=0.2]{};

\node (mask_anno) [ellipses, fill=white, minimum width=8ex, minimum height=3ex, rotate=0]{};
\node (rect_box_anno)[block, thick,draw=red,opacity=1.0, sharp corners, minimum width=10ex, minimum height=4ex,xshift=0.2ex,yshift=-1.0ex]{};

\node (rect_bbox_anno)[block, thick,draw=green,opacity=1.0, sharp corners, minimum width=8ex, minimum height=3ex]{};
\end{tikzpicture}
};
\node (node_equal2)[below of=node_equal1, yshift=-4ex,scale=\nodescale]{
\begin{tikzpicture}
\node (rect_)[block,fill=black,sharp corners, minimum width=\linewidth, minimum height=0.8*\linewidth,scale=0.2]{};

\node (mask_anno) [ellipses, fill=white, minimum width=8ex, minimum height=3ex, rotate=0]{};

\node (rect_box_anno)[block,thick,fill=black,opacity=1.0, sharp corners, minimum width=10ex, minimum height=2ex,xshift=0.0ex,yshift=2.0ex]{};

\node (rect_box_anno)[block,thick,draw=cyan,opacity=1.0, sharp corners, minimum width=8ex, minimum height=2.3ex,xshift=0.01ex,yshift=-0.3ex]{};

\end{tikzpicture}
};

\node (node_greater1)[scale=\nodescale, xshift=25ex]{
\begin{tikzpicture}
\node (rect)[block,fill=black,sharp corners, minimum width=\linewidth, minimum height=0.8*\linewidth,scale=0.2]{};

\node (mask_anno) [ellipses, fill=white, minimum width=8ex, minimum height=3ex, rotate=0]{};

\node (rect_box_err)[block,thick,fill=black,opacity=1.0, sharp corners, minimum width=2ex, minimum height=5ex,xshift=-4ex]{};

\node (rect_bbox_anno)[block,thick,draw=green,opacity=1.0, sharp corners, minimum width=6.8ex, minimum height=3ex, xshift=0.45ex]{};

\node (rect_box_anno)[block,thick,draw=red,opacity=1.0, sharp corners, minimum width=9ex, minimum height=4ex,xshift=0.5ex,yshift=-1ex]{};

\end{tikzpicture}
};
\node (node_greater2)[below of=node_greater1, yshift=-4ex,scale=\nodescale]{
\begin{tikzpicture}

\node (rect_)[block,fill=black,sharp corners, minimum width=\linewidth, minimum height=0.8*\linewidth,scale=0.2]{};
\node (mask_anno) [ellipses, fill=white, minimum width=8ex, minimum height=3ex, rotate=0]{};

\node (rect_box_err)[block,thick,fill=black,opacity=1.0, sharp corners, minimum width=2ex, minimum height=5ex,xshift=-4ex]{};

\node (rect_bbox_anno)[block,thick,draw=cyan,opacity=1.0, sharp corners, minimum width=6.8ex, minimum height=3ex, xshift=0.45ex]{};

\end{tikzpicture}
};

\node (less_text) [below of =node_less2]{ $\mathbf{p_{m} < p_{b}}$};
\node (equal_text) [below of =node_equal2]{ $\mathbf{p_{m} = p_{b}}$};
\node (greater_text) [below of =node_greater2]{ $\mathbf{p_{m} > p_{b}}$};

\node (rect1) at (node_less1) [draw=white!70!black,minimum width=10ex, minimum height=22.5ex, rounded corners=0.6mm,line width=0.1pt,yshift=-7ex]{};

\node (rect2) at (node_equal1) [draw=white!70!black,minimum width=10ex, minimum height=22.5ex, rounded corners=0.6mm,line width=0.1pt,yshift=-7ex]{};

\node (rect3) at (node_greater1) [draw=white!70!black,minimum width=10ex, minimum height=22.5ex, rounded corners=0.6mm,line width=0.1pt,yshift=-7ex]{};
\draw [->, very thick] (node_less1) -- (node_less2);
\draw [->, very thick] (node_equal1) -- (node_equal2);
\draw [->, very thick] (node_greater1) -- (node_greater2);

\end{tikzpicture}
};
\end{tikzpicture}

\label{fig_priorities_auto_anno}
}

\newcommand{\greenrect}{\raisebox{0.12ex}{\tikz{\node (rect1) [rectangle,draw = green,scale=0.5,line width=1.1pt]{};}}}
\newcommand{\redrect}{\raisebox{0.12ex}{\tikz{\node (rect1) [rectangle,draw = red,scale=0.5,line width=1.1pt]{};}}}
\newcommand{\cyanrect}{\raisebox{0.12ex}{\tikz{\node (rect1) [rectangle,draw = cyan,scale=0.5,line width=1.1pt]{};}}}

\caption{(a) Single instance images, (b) sample predicted mask (\protect\greenrect) and box (\protect\redrect), and the combined output (\protect\cyanrect) for various cases}

\vspace*{-0.3ex}
\end{figure}

Our motivation to develop this technique lies behind the need of pixel-wise ground truth masks in order to train the child network for semantic segmentation. Typically, it requires $1$-$2$ hours of manual effort to generate masks for roughly $60$ images similar to Fig. \ref{fig_dataset}. For this reason, manual generation of such masks in $45$ minutes was not feasible in ARC'$17$.
%Typically, pixelwise and bounding box annotations are required in order to train CNNs for the task of semantic segmentation and bounding box regression. In our case, such annotations of novel items were required in order to train the child network for the task of semantic segmentation. 
%In general, such annotations are generated manually and for this reason, it was difficult to achieve that for the new items only in $45$ minutes.
In order to automate the labeling process, the traditional approaches such as background subtraction and depth segmentation are not sufficient because they involve a number of hyper-parameters such as color difference thresholds, depth threshold, and kernel size of morphological operations. It becomes quite challenging to tune these parameters for uncontrolled scenarios such as lighting variations, reflections and the items which have transparent or wired mesh surfaces, or appears similar to a given background.%surface reflections,  transparent and wired mesh items and also, the items which appears similar to the background color.
% In addition, deploying multiple depth sensors to capture multiple views in a single shot would increase the system cost. 
%
\par
Hence, we devise an alternative approach which addresses all of the above issues at once and produces high quality mask and bounding box annotations in real time\footnote{depending on hardware configuration}, with no manual supervision. The approach is inspired by the outstanding capabilities of humans to detect and segment novel objects. It encourages us to assume that a part of human vision system possibly functions in a class agnostic manner, i.e. detection and segmentation are done irrespective of the category. To replicate this behavior, we collect single instance images (Fig. \ref{fig_dataset}) and use them in order to train two CNNs: one for semantic segmentation and another for bounding box regression. We refer these CNNs as tutors. The detailed procedure to train the tutors is given in Sec. \ref{subsec_experiments}.
\par
\begin{figure}
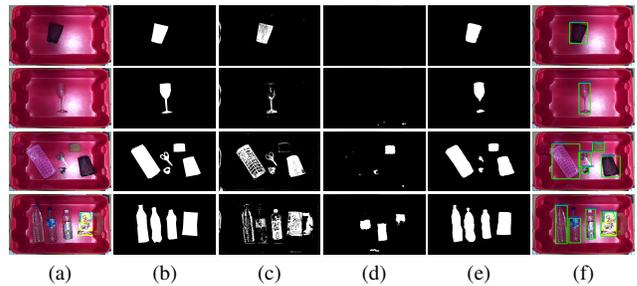

\centering

\FPeval{\width}{8.5ex}
\FPeval{\height}{5ex}
\FPeval{\shift}{0.3ex}

\begin{tikzpicture}

\foreach \i in {2,...,5}
\node (node_\i) [rectangle, xshift=0 * (\width +\shift),yshift=-\i * (\height+\shift)]{
\includegraphics[width=\width,height=\height]{images/auto_anno/foscam_\i.png}};

\foreach \i in {2,...,5}
\node (node_\i) [rectangle, xshift=1 * (\width +\shift),yshift=-\i * (\height+\shift)]{
\includegraphics[width=\width,height=\height]{images/auto_anno/gnd_truth_\i.png}};

\foreach \i in {2,...,5}
\node (node_\i) [rectangle, xshift=2 * (\width +\shift),yshift=-\i * (\height+\shift)]{
\includegraphics[width=\width,height=\height]{images/auto_anno/bckgnd_mask_\i.png}};

\foreach \i in {2,...,5}
\node (node_\i) [rectangle, xshift=3 * (\width +\shift),yshift=-\i * (\height+\shift)]{
\includegraphics[width=\width,height=\height]{images/auto_anno/mask_3d_\i.png}};

\foreach \i in {2,...,5}
\node (node_\i) [rectangle, xshift=4 * (\width +\shift),yshift=-\i * (\height+\shift)]{
\includegraphics[width=\width,height=\height]{images/auto_anno/foscam_deep_mask_\i.png}};

\foreach \i in {2,...,5}
\node (node_\i) [rectangle, xshift=5 * (\width +\shift),yshift=-\i * (\height+\shift)]{
\includegraphics[width=\width,height=\height]{images/auto_anno/foscam_box_\i.png}};

\FPeval{\subfloatcaptiony}{0-5.8}

\node (node_1) [rectangle, xshift=0 * (\width +\shift),yshift=\subfloatcaptiony * (\height+\shift)]{\footnotesize (a)};
\node (node_2) [rectangle, xshift=1 * (\width +\shift),yshift=\subfloatcaptiony * (\height+\shift)]{\footnotesize (b)};
\node (node_3) [rectangle, xshift=2 * (\width +\shift),yshift=\subfloatcaptiony * (\height+\shift)]{\footnotesize (c)};
\node (node_4) [rectangle, xshift=3 * (\width +\shift),yshift=\subfloatcaptiony * (\height+\shift)]{\footnotesize (d)};
\node (node_5) [rectangle, xshift=4 * (\width +\shift),yshift=\subfloatcaptiony * (\height+\shift)]{\footnotesize (e)};
\node (node_6) [rectangle, xshift=5 * (\width +\shift),yshift=\subfloatcaptiony * (\height+\shift)]{\footnotesize (f)};

\end{tikzpicture}
\caption{(a) Input image, (b) ground truth mask. Mask generation using (c) background subtraction, (d) depth information. (e) Semi supervised labeling for mask, and (f) box labeling.}
\label{fig_auto_anno_results}
\end{figure}

The tutors can successfully detect and segment transparent objects as long as they are differentiable in the image by visual inspection. We exploit comprehensive data augmentation (Sec. \ref{experiments}) and the property of CNNs to learn contextual information in order to enable the tutors to handle background color changes as well as the cases of color similarity between the object and the background. The approach doesn't require any background modeling unlike the background and depth subtraction. For this reason, it can be employed in the warehouses or any other relevant area without a need to model the background each time. The effectiveness of the tutors can be examined by Fig. \ref{fig_auto_anno_results} which shows qualitative results of the traditional approaches and the tutors to annotate variety of items i.e. transparent (rows $2$-$5$), mesh (row $1$).
\par
Further, we combine the predicted mask and the box in order to produce more robust ground truths, especially when either or both of the annotations are not accurate (Fig. \ref{fig_priorities_auto_anno}). Let $\mathbf{p_{m},p_{b}}$ be the priorities of mask and box annotations. It must be noticed that the priorities are merely fixed numbers e.g. $\mathbf{\{p_{m}, p_{b}\} = \{1,0\},\{0,1\}}$ or $\mathbf{\{1,1\}}$. These values repersents three configurations given below.
\begin{enumerate}
\item  if $\mathbf{p_{m} < p_{b} }$, final mask is given by the $box_{b}$.
\item  if $\mathbf{p_{m} = p_{b} }$, it is given by $box_{m} \cap box_{b}$, and
\item  if $\mathbf{p_{m} > p_{b} }$, it is given by $box_{m}$.
\end{enumerate}
\par
Where $box_{m}$ be the bounding box around the predicted mask and $box_{b}$ be a predicted box annotation respectively. Later, the obtained class agnostic annotations are assigned a physical label such as a box, crayons or bottle etc., provided by a human. For this reason, we call it semi-supervised labeling as the mask or box is generated by the tutor while a meaningful label is provided by human. 
\subsection{\textbf{Occlusion Aware Scene Synthesis}}
\begin{figure}[t]
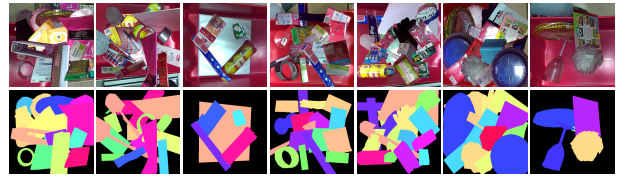

\vspace*{0.3ex}

\centering
\begin{tikzpicture}
\FPeval{\width}{7ex}
\FPeval{\height}{7ex}

\foreach \i in {1,...,7}
\node (node_\i) [rectangle, xshift= \i * (\width + 0.3ex)]{\includegraphics[width=\width,height=\height]{images/clutter_\i.jpg}};

\foreach \i in {1,...,7}
\node (node_\i_) [rectangle, xshift= \i * (\width + 0.3ex),yshift=-\height - 0.3ex]{
\includegraphics[width=\width,height=\height]{images/clutter_coloured_mask_\i.png}};

%\node [label={[rotate=90]right:\textbf{Synthetic clutter}}, xshift=5ex, yshift=-8.5ex]{};
%\node [label={[rotate=90]right:\textbf{Mask}}, xshift=5ex, yshift=-19ex]{};

\end{tikzpicture}

\caption{Generated synthetic clutter and color coded labels}
\label{fig_synthetic_clutter}

\vspace*{-0.3ex}

\end{figure}

Typically, the performance of CNNs heavily relies on the nature of a dataset. Therefore, a CNN gets biased when trained only for single instance images (Fig. \ref{fig_dataset}) and exhibits poor performance on the cluttered scenes. Hence, cluttered images must be a part of the dataset in order to improve the performance. However, the cost and time to manually annotate such images, is directly related to the number of classes and instances per image. For this reason, the collection and manual labeling of such images is infeasible in short durations, similar to ARC'$17$. Hence, we develop an effective occlusion aware clutter synthesizing technique which can generate large number of realistic multi-class cluttered scenes (Fig. \ref{fig_synthetic_clutter}) along with ground-truth labels from only a few samples of the images similar to Fig. \ref{fig_dataset}.
\par
Let there be a set of $K$ classes and each class $c \in K$ has a total of $I_c$ images. To generate synthetic clutter, an image is randomly chosen from the dataset or sample background images (if any). We refer it as a \texttt{base-image}. The base-image is then partitioned into a grid of size $M \times M$. The value of $M$ is selected from a set of predefined values and governs the clutter level in the output image. For ARC'$17$, we use low, medium and high clutter having grid size of $3 \times 3,~ 4 \times 4,~ 5 \times 5$ respectively. For each grid center, a category $c \in K$ and image $I \in I_{c}$ is selected randomly. The pixels belonging to class $c$ in the image $I$ are rendered onto the base-image such that center of the object to be rendered coincides with the grid center. In this process, a corresponding ground truth image is also updated parallely.
\par
The above naive way of generating synthetic clutter often leads to situations where a significant part of an item is occluded and the remaining visible part doesn't contain enough contextual information. Such parts are redundant and might resemble with other objects and degrades the segmentation performance. Thus, we eliminate the object or its part in the clutter, which is below a visibility threshold. It is done by keeping a track of number of pixel of an object before copying and after clutter generation. The works \cite{nimbro2017} and \cite{cutpaste} are an example of coexisting work during the time-line of ARC'$17$, however, our approach is entirely different from them.

\subsection{\textbf{Customized CNN Head}}
\label{subsec_customized_nw}
Generation of unique and separable features has been one of the aims of the machine learning algorithms since separablility in the features improves the classification accuracies. Hence, we reduce the challenge of quick learning, to embed uniqueness in the features at the deeper layers of a CNN. To achieve this, we develop a CNN head based on a feature pyramid module \cite{fpn} and insert a few branches to learn contextual information. Fig. \ref{fig_customized_neural_links} shows the child CNN with ResNet-$50$ \cite{resnet} as its backbone along with the proposed head. To embed uniqueness, we merge features from the last four stages of the ResNet-$50$ i.e. conv$2\_3$, conv$3\_4$, conv$4\_6$, and conv$5\_3$. The first stage conv$1\_3$ avoided due to overly large GPU memory footprints. The feature merging is achieved by three kinds of blocks discussed below.
\par
\subsubsection{\textbf{Feature Smoothing Block}} FSB is a stack of $1 \times 1$ Convolution - Batch\_Normalization - ReLU. This block reduces the dimensionality or smoothen the input features.
\par
\subsubsection{\textbf{Feature Interpolator Block}} At each stage of ResNet-$50$, the spatial size of feature is reduced by half in order to increase the receptive field. Hence, output features of the deeper layers need to be upsampled in order to merge them with the features produced by shallower layers. FIB perform this operation by using bilinear interpolation. Although, a deconvolution layer can also be used for this purpose, we avoid its use because of extra learnable parameters.
\par
\subsubsection{\textbf{Context Extraction Block}} 
\begin{figure}[t]
\centering
\subfloat[]
{

\begin{tikzpicture}

\FPeval{\txthgt}{0.4}
\FPeval{\radiicornr}{0.15}

\node (ex) [scale = 0.95]{
\begin{tikzpicture}[node distance=5.0ex,scale=0.5]

\node (resnet50) [rectangle, fill=cyan!90!magenta,label={[above]\tiny  Resnet-$50$},minimum width=6.5ex,minimum height=17ex,yshift=-7.0ex]{};

\node (stage1) [rectangle, fill=black!10!white,text height=\txthgt ex,rounded corners=\radiicornr mm]{\tiny Conv$1\_3$};
\node (stage2) [rectangle, fill=black!10!white, below of=stage1,yshift=1.5ex,text height=\txthgt ex,rounded corners=\radiicornr mm]{\tiny Conv$2\_3$};
\node (stage3) [rectangle, fill=black!10!white, below of=stage2,yshift=1.5ex,text height=\txthgt ex,rounded corners=\radiicornr mm]{\tiny Conv$3\_4$};
\node (stage4) [rectangle, fill=black!10!white, below of=stage3,yshift=1.5ex,text height=\txthgt ex,rounded corners=\radiicornr mm]{\tiny Conv$4\_6$};
\node (stage5) [rectangle, fill=black!10!white, below of=stage4,yshift=1.5ex,text height=\txthgt ex,rounded corners=\radiicornr mm]{\tiny Conv$5\_3$};

\node (fib1) [rectangle, fill=skyblue!100!white, below of=stage5,yshift=1.5ex,text height=\txthgt ex,rounded corners=\radiicornr mm]{\tiny FIB};
\node (fmb1) [rectangle, fill=gray!60!white, left of= fib1, below of=stage5, xshift=1ex,yshift=1.5ex,text height=\txthgt ex,rounded corners=\radiicornr mm]{\tiny $+$};
\node (fsb1) [rectangle, fill=violet!40!white, below of=fmb1,yshift=1.5ex,text height=\txthgt ex,rounded corners=\radiicornr mm]{\tiny FSB};

\node (fib2) [rectangle, fill=skyblue!100!white, below of=fsb1,yshift=1.5ex,text height=\txthgt ex,rounded corners=\radiicornr mm]{\tiny FIB};
\node (fmb2) [rectangle, fill=gray!60!white, left of=fib2, xshift=0.9ex,text height=\txthgt ex,rounded corners=\radiicornr mm]{\tiny $+$};
\node (fsb2) [rectangle, fill=violet!40!white, below of=fmb2,yshift=1.5ex,text height=\txthgt ex,rounded corners=\radiicornr mm]{\tiny FSB};

\node (fib3) [rectangle, fill=skyblue!100!white, below of=fsb2,yshift=1.5ex,text height=\txthgt ex,rounded corners=\radiicornr mm]{\tiny FIB};
\node (fmb3) [rectangle, fill=gray!60!white, left of=fib3, xshift=0.9ex,text height=\txthgt ex,rounded corners=\radiicornr mm]{\tiny $+$};
\node (fsb3) [rectangle, fill=violet!40!white, below of=fmb3,yshift=1.5ex,text height=\txthgt ex,rounded corners=\radiicornr mm]{\tiny FSB};

\node (fib4) [rectangle, fill=skyblue!100!white, right of=fib3, xshift=6.5ex,yshift=6.3ex,text height=\txthgt ex,rounded corners=\radiicornr mm]{\tiny FIB};
\node (fsb4) [rectangle, fill=violet!40!white, right of=fsb3, xshift=10.6ex ,yshift=6.3ex,text height=\txthgt ex,rounded corners=\radiicornr mm]{\tiny FSB};

\node (fib5) [rectangle, fill=skyblue!100!white, right of=fib4, xshift=-1.3ex,text height=\txthgt ex,rounded corners=\radiicornr mm]{\tiny FIB};
\node (fsb5) [rectangle, fill=violet!40!white, right of=fsb4, xshift=-1.3ex,text height=\txthgt ex ,rounded corners=\radiicornr mm]{\tiny FSB};

\node (fib6) [rectangle, fill=skyblue!100!white, right of=fib5, xshift=-1.3ex,text height=\txthgt ex,rounded corners=\radiicornr mm]{\tiny FIB};
\node (fsb6) [rectangle, fill=violet!40!white, right of=fsb5, xshift=-1.3ex,text height=\txthgt ex,rounded corners=\radiicornr mm]{\tiny FSB};

\node (fib7) [rectangle, fill=skyblue!100!white, right of=fib6, xshift=-1.3ex,text height=\txthgt ex,rounded corners=\radiicornr mm]{\tiny FIB};
\node (fsb7) [rectangle, fill=violet!40!white, right of=fsb6, xshift=-1.3ex,text height=\txthgt ex,rounded corners=\radiicornr mm]{\tiny FSB};

\node (flb) [rectangle, fill=purple!50!white, below of=fib1, xshift = 1ex, yshift=-12.5ex,text height=\txthgt ex,rounded corners=\radiicornr mm]{\tiny CEB};
\node (fsb8) [rectangle, fill=violet!40!white, below of=flb,yshift=1.5ex,text height=\txthgt ex,rounded corners=\radiicornr mm]{\tiny FSB};

\draw [arrow] (stage1) -- (stage2);
\draw [arrow] (stage2) -- (stage3);
\draw [arrow] (stage3) -- (stage4);
\draw [arrow] (stage4) -- (stage5);

\draw [arrow] (stage5) -- (fib1);
\draw [arrow] (stage4) -| (fmb1);
\draw [arrow] (fib1) -- (fmb1);
\draw [arrow] (fmb1) -- (fsb1);

\draw [arrow] (fsb1) -- (fib2);
\draw [arrow] (stage3) -| (fmb2);
\draw [arrow] (fib2) -- (fmb2);
\draw [arrow] (fmb2) -- (fsb2);

\draw [arrow] (fsb2) -- (fib3);
\draw [arrow] (stage2) -| (fmb3);
\draw [arrow] (fib3) -- (fmb3);
\draw [arrow] (fmb3) -- (fsb3);

\draw [arrow] (stage5) -| (fib4);
\draw [arrow] (fib4) -- (fsb4);

\draw [arrow] (stage4) -| (fib5);
\draw [arrow] (fib5) -- (fsb5);

\draw [arrow] (stage3) -| (fib6);
\draw [arrow] (fib6) -- (fsb6);

\draw [arrow] (stage2) -| (fib7);
\draw [arrow] (fib7) -- (fsb7);

\draw [arrow] (fsb3) |- (flb);
\draw [arrow] (fsb4) -| (flb);
\draw [arrow] (fsb5) |- ($(flb.east) - (0, -1ex)$);
\draw [arrow] (fsb6) |- (flb);
\draw [arrow] (fsb7) |- ($(flb.east) - (0, 1ex)$);

\draw [arrow] (flb) -- (fsb8);

\end{tikzpicture}
};
\end{tikzpicture}

\label{fig_customized_neural_links}
}
\hspace*{0.5ex}
\subfloat[]
{

\FPeval{\radiicornr}{0.15}

\begin{tikzpicture}
\FPeval{\width}{16}
\FPeval{\height}{12}

\FPeval{\imwidth}{2}
\FPeval{\imheight}{2}

\FPeval{\rotimwidth}{15.4}
\FPeval{\rotimheight}{0.75*\rotimwidth}

\node (rot_plat) [rectangle, minimum width=\width ex,minimum height=\height ex, xshift=0 ex]{\includegraphics[width=\rotimwidth ex,height=\rotimheight ex]{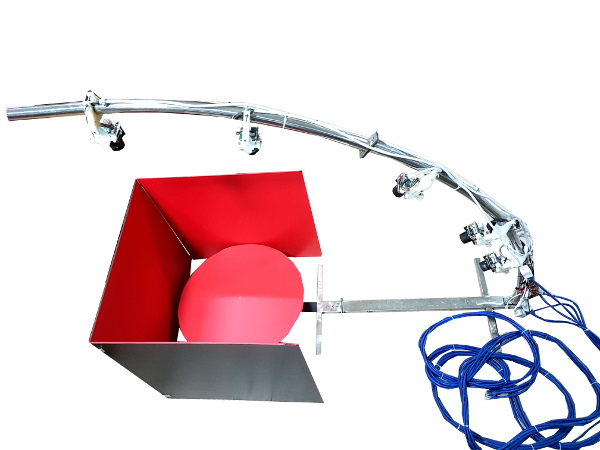}};

\node (rect_nodplat) at(rot_plat) [draw=white!70!black,minimum width = 17ex, minimum height=11ex, rounded corners=0.6mm,line width=0.1pt, yshift=-1ex]{};

\node (rot_plat_text) [rectangle, below of=rot_plat, xshift=-0.5ex,yshift= 0.5ex]{\tiny Rotating platform};

\node (online_images) [block, below of=rot_plat, fill=red!60!white,minimum width=\width ex,minimum height=0.1*\height ex, xshift=0 ex,yshift=-5ex,rounded corners=\radiicornr mm]{\tiny Image Aquisition};

\foreach \i in {1,...,3}
{
\FPeval{\del}{\i*\imwidth}
\FPeval{\imname}{clip(\i+35)}
\node (online_image_\i) [rectangle,below of=online_images, xshift=-\del ex -0.5, yshift=3.5 ex]{\includegraphics[width=\imwidth ex,height=\imheight ex]{images/\imname.png}};
}

\node (auto_anno) [block, fill=red!60!white, below of=online_images,minimum width=\width ex,minimum height=0.1*\height ex, yshift=0.5 ex,rounded corners=\radiicornr mm]{\tiny Tutor};

\foreach \i in {1,...,3}
{
\FPeval{\del}{\i*\imwidth}
\FPeval{\nodnum}{clip(\i+4)}
\FPeval{\imname}{clip(\i+35)}
\node (online_image_\nodnum) [rectangle,right of=auto_anno, xshift=-\del ex -6.5 ex, yshift=-2.8 ex]{\includegraphics[width=\imwidth ex,height=\imheight ex]{images/\imname.png}};
}

\foreach \i in {1,...,3}
{
\FPeval{\del}{\i*\imwidth}
\FPeval{\imname}{clip(\i+35)}
\node (online_seg_\i) [rectangle,right of=auto_anno, xshift=-\del ex + 1.9 ex, yshift= -2.8 ex]{\includegraphics[width=\imwidth ex,height=\imheight ex]{images/coloured_mask_\imname.png}};
}

\node (synthetic_clutter) [block, fill=red!60!white,below of=auto_anno, minimum width=\width ex,minimum height=0.1*\height ex, yshift=0.5 ex,rounded corners=\radiicornr mm]{\tiny Synthetic cluttering};

\foreach \i in {1,...,3}
{
\FPeval{\del}{\i*\imwidth}
\node (clutter_image_\i) [rectangle,right of=synthetic_clutter, xshift=-\del ex - 6.5 ex, yshift=-2.9 ex]{\includegraphics[width=\imwidth ex,height=\imheight ex]{images/clutter_\i.jpg}};
}

\foreach \i in {1,...,3}
{
\FPeval{\del}{\i*\imwidth}
\node (clutter_seg_\i) [rectangle,right of=synthetic_clutter, xshift=-\del ex + 1.9 ex, yshift= -2.9 ex]{\includegraphics[width=\imwidth ex,height=\imheight ex]{images/clutter_coloured_mask_\i.png}};
}

\node (training) [block, fill=red!60!white, below of=synthetic_clutter,minimum width=\width ex,minimum height=0.1*\height ex, yshift=0.5 ex,rounded corners=\radiicornr mm]{\tiny Child};

\node (human_input) [rectangle,above of=auto_anno,minimum height=0.1*\height ex, yshift=-3ex, xshift=4ex]{\tiny class label};

\node () [below of=training, yshift=1ex]{};

\draw [arrow] (online_images) -- (auto_anno);
\draw [arrow] (auto_anno) -- (synthetic_clutter);
\draw [arrow] (synthetic_clutter) -- (training);
\draw [arrow] (human_input.south) -- ($(auto_anno.north)+(4 ex,0)$);
\draw [<->] (rot_plat.south) -- (online_images.north);

\end{tikzpicture}
%\end{turn}
\label{fig_state_transition}
}
\caption{(a) Child with the customized CNN head, (b) state transition in the proposed online learning system}
\end{figure}

Utility of this block is to increase feature separability by gathering various levels of context. This block performs a concatenation operation which combines diverse features from earlier stages to embed uniqueness in features (Fig. \ref{fig_customized_neural_links}).
\subsection{\textbf{Instance Detection and Segmentation}}
The semantic segmentation approach assigns a common label to all instances of a class and doesn't differentiate between them. In ARC'$17$, all items had exactly one instance and therefore, semantic segmentation was sufficient for recognition purpose by treating each item as a different category. However, multiple instances are quite common practically. In such cases, The Mask-RCNN \cite{maskrcnn} can not be employed because the box regression is hard to achieve in short durations ($\sim30$ minutes) of
training as compared to semantic segmentation and is also computationally slower. Hence, we extend our system to perform instance detection and segmentation while adding minimal computational overhead.
\par
To achieve this, first, pixel-wise semantic labels are for an input image are obtained by the child. For each item class, pixel-wise masked images are computed using the predicted labels i.e. $n$ masked images for $n$ classes. All such images are then fed to the tutor network which predicts bounding boxes for all instances of the class, similar to Fig. \ref{fig_auto_anno_results}f. The predicted boxes serves as the rectangular boundaries for each instance while the labels predicted by the child serves as the segmentation mask (Fig. \ref{fig_comptetition_runs}).
\section{Online Learning}
\label{implementation}
\label{subsec_implementation}
%
%Each of the components described above had played a major role in our performance during ARC'$17$.
In an offline learning system, first, data is collected, labeled and then learning is performed. This approach, in the warehouse automation, is not feasible as novel items keep on arriving frequently and it becomes difficult to incorporate them into the existing learned models. For example, let $20$ items have already been learned by the system. In order to learn an additional item, an offline system would require to first generate the ground truth and restart the training process. Hence, we improve our system to learn in an online manner\footnote{not to be confused with reinforcement learning}, i.e. the system can acquire the images, label them, generate synthetic scenes and perform the learning process, altogether. Our system doesn't encounter catastrophic forgetting \cite{catastrophic} of already learned items due to inclusion of all the available items through the synthesized clutter. 
\par
To realize a practical online learning system, we interconnect all the system components and deploy them on a Multi-GPU server having $8\times$NVIDIA Geforce GTX-1080Ti, $256$ GB RAM and $2\times$Intel Xeon(R)-E5-2683-v4 CPUs, each having $16$ physical cores. We use Caffe \cite{caffe} with C++ architecture and extensively modify its core internals to support the proposed online learning functionality. Our implementation of the learning process (Fig. \ref{fig_state_transition}) is complex synchronization of multiple threads and can be divided mainly into four stages.
\par
\subsubsection{\textbf{Image Acquisition}}
The image acquisition process is facilitated by a rotating platform, equipped with $5\times$FOSCAM FI9903P HD RGB LAN cameras mounted with different viewing angle. We divide one revolution ($\sim10~sec$) into $12$ parts which results in $12 \times 5 = 60 $ images per revolution. For each object, we repeat the process for two revolutions, in order to capture all views of an item. With these settings, image can be collected at a rate of $\sim360$ images per minute (Table \ref{tab_data_aqs}). Alternative to the platform, our system can also be taught by showing an item to the camera by hand.
\par
\subsubsection{\textbf{Ground Truth Generation}}
The acquired images are sent to the tutor to obtain corresponding class agnostic mask and box annotations. These annotations are combined with priorities $\mathbf{p_{m} = p_{b}}$ (Fig. \ref{fig_priorities_auto_anno}) to obtain a final mask which is later filled with a numeric-id (class label), either provided by a human or computerized file storages. Due to real time speed, the ground truth is generated as soon as the images are acquired. To accomplish the same task, it would require approximately $1-2$ hours of human effort. Our system is capable of generating fully annotated images at $~10-15$ FPS (Table \ref{tab_data_aqs}). This speed can be increased by using high-speed cameras to avoid motion blur occurring due to rotating platform.
\par
\subsubsection{\textbf{Clutter Synthesis}}
The images and corresponding ground truths obtained from the previous step are used to synthesize cluttered scenes. A total of $24$ threads are responsible and all of them remain live as long as the training runs. The clutter is generated at a rate of $\sim5\times24=120$ FPS and in this process, most of the time is spent in the image decoding. It can be reduced by prefetching all available images in RAM and continuously discarding the clutter images which have been used in the training process. This increases scene synthesis throughput (Table \ref{tab_clutter}).
\par
\subsubsection{\textbf{Child Learning}}
The child CNN is replicated across all the available GPUs. The child is fed with both the single-class (obtained from platform) and multi-class images (cluttered scenes). The selection between single-class and multi-class images is done at random with a probability ratio of $1:3$. The timing performance of the child learning is provided in the Table \ref{tab_data_aqs}. On the specified server, It can learn at $32$ FPS i.e. $4\times8$ (Batch size$\times$GPUs)

\section{Experiments}
\label{experiments}
\label{subsec_experiments}
\begin{table}[t]
\vspace*{0.9ex}

\centering
\caption{\footnotesize Timing analysis of online learning process}
\label{tab_data_aqs}
\arrayrulecolor{white!60!black}

\begin{tabular}{c|c|c|c|c}
\hline
%Time per revolution (sec) & $5$ & $10$ & $15$ & $20$ \\ \hline
%\multicolumn{5}{|c|}{Data Aquisition} \\ \hline
\shortstack{Angle resolution} & $60^{\circ}$ & $30^{\circ}$ & $20^{\circ}$ & $10^{\circ}$ \\ \hline
Acquired Images (per rev $\sim$$10$ sec) & $30$ & $60$ & $108$ & $120$ \\ \hline
Frame grabbing (per image) & \multicolumn{4}{c}{$40$ ms ($25$ FPS)} \\ 
Communication overhead ($5$ cameras)& \multicolumn{4}{c}{$90$ ms ($62$ FPS)} \\ \hline
Tutor-Box annotation (per image) & \multicolumn{4}{c}{$$ $70$ ms ($14$ FPS)}\\ 
Turor-Mask annotation (per image) & \multicolumn{4}{c}{$$ $120$ ms ($8$ FPS)}\\ \hline
Child learning time (per image) & \multicolumn{4}{c}{$270$ ms ($4$ FPS)} \\ \hline

\end{tabular}

\end{table}
\begin{table}[t]
\centering
\caption{\footnotesize Timing analysis of occlusion aware synthetic clutter}
\label{tab_clutter}

\arrayrulecolor{white!60!black}

\begin{tabular}{c|c|c|c|c|c}
\hline
\multirow{2}{*}{Grid size} & \multicolumn{3}{c|}{Time per operation (ms)} & \multicolumn{2}{c}{Total time (ms)} \\ \cline{2-6}
 & \makecell{image \\ decoding} & \makecell{object\\ transfer} & \makecell{visibility \\ check} & \makecell{disk\\ read} & \makecell{prefetch \\ all in RAM} \\ \hline
 $3\times 3$ & $180$ & $63$  & $69$  & $320$ & $135$ \\ %\hline
 $4\times 4$ & $272$ & $112$ & $108$ & $503$ & $220$ \\ %\hline
 $5\times 5$ & $475$ & $150$ & $140$ & $767$ & $290$ \\ \hline

\end{tabular}

\end{table}
\begin{table}[t]
\vspace*{0.9ex}

\centering
\caption{\footnotesize Performance analysis of the Tutors}
\label{tab_autoanno}

\arrayrulecolor{white!60!black}

\begin{tabular}{c c c c c c c}
\hline

\multicolumn{5}{c}{Augmentation} & \multirow{2}{*}{\makecell{Mask\\ mIoU}} & \multirow{2}{*}{\makecell{Box\\ mIoU}} \\ \cline{1-5}

colour & scale & mirror & blur & rotate & & \\ \hline
 \xmark  & \xmark & \xmark & \xmark & \xmark & \multicolumn{1}{|c}{$45.9$} & $33.1$\\ 
 \cmark  &  &  &  &  & \multicolumn{1}{|c}{$80.9$} & $77.3$ \\ 
 \cmark  & \cmark &  &  &  & \multicolumn{1}{|c}{$89.4$} & $81.6$\\ 
 \cmark  & \cmark & \cmark &  &  & \multicolumn{1}{|c}{$90.2$} & $84.9$\\ 
 \cmark  & \cmark & \cmark & \cmark &  & \multicolumn{1}{|c}{$92.7$} & $86.8$\\ 
 \cmark  & \cmark & \cmark & \cmark & \cmark & \multicolumn{1}{|c}{\textcolor{blue}{$94.6$}} & \textcolor{blue}{$87.6$} \\ \hline

\end{tabular}
\end{table}
\begin{table}[t]
\centering
\caption{\footnotesize Performance analysis of the Child and state-of-the-art}

\label{tab_segmentation_perfm}

\arrayrulecolor{white!60!black}

\begin{tabular}{c c c c c c c}
\hline

\multicolumn{3}{c}{Clutter level} & \multicolumn{4}{|c}{\makecell{mIoU}}  \\  \hline

\multirow{2}{*}{$3\times 3$} & \multirow{2}{*}{$4\times 4$} & \multirow{2}{*}{$5\times 5$} & \multicolumn{2}{|c}{$\eta=0.01$} & \multicolumn{2}{|c}{$\eta=0.001$} \\ \cline{4-7} 

 & & & \multicolumn{1}{|c}{PSPNet} & \multicolumn{1}{|c}{Child} & \multicolumn{1}{|c}{PSPNet} & \multicolumn{1}{|c}{Child} \\ \hline

\xmark  & \xmark & \xmark & \multicolumn{1}{|c}{$51.2$} & \multicolumn{1}{c}{$51.4$} & \multicolumn{1}{|c}{$59.9$} & \multicolumn{1}{c}{$77.8$}    \\

 \cmark  &  &  & \multicolumn{1}{|c}{$52.3$} & \multicolumn{1}{c}{$55.7$} & \multicolumn{1}{|c}{$62.9$} & \multicolumn{1}{c}{79.5}    \\

 \cmark  & \cmark &  & \multicolumn{1}{|c}{$58.1$} & \multicolumn{1}{c}{$65.3$} & \multicolumn{1}{|c}{$73.7$} & \multicolumn{1}{c}{$84.1$}    \\

 \cmark  & \cmark & \cmark & \multicolumn{1}{|c}{$62.8$} & \multicolumn{1}{c}{\textcolor{blue}{$90.2$}} & \multicolumn{1}{|c}{$84.8$} & \multicolumn{1}{c}{\textcolor{blue}{$87.2$}}    \\
\hline

\end{tabular}
\end{table}
%
%
%\begin{figure}[h]
%\caption{Qualitative results of semi supervised labelling of novel items}
%\label{fig_semi_supervised_labelling_eval}
%\end{figure}
%
%
%
%We evaluate all the system components in order to validate their practical importance. First, we report quantitative as well as qualitative analysis of each component and their effect on overall learning. Later, we discuss the overall system in the context of ARC'$17$. All the experiments were performed on the multi-GPU server mentioned previously.
%
\subsection{\textbf{Dataset}}
We collect $12000$ single instance images of $40$ items provided by Amazon a priori, referred as \texttt{known-set}. Then we manually generate their mask and box annotations. During ARC'$17$, each task had a \texttt{competition-set} comprising of known and novel items divided equally. The stow-task $20$, pick-task $32$ and final stow-pick task had $32$ items. The images of novel items were collected during $45$ minutes. As mentioned previously, $120$ images are obtained for two revolution of the platform and each was rotated by $-10^{\circ}$ and $10^{\circ}$ in order to approximately match the already collected $300$ images per known item.
\subsection{\textbf{Semi Supervised Labeling}}
We split the set of $12000$ images into two sets of $8000$ and $4000$ i.e. $200$ train and $100$ test images per item. We train the network architecture (Fig. \ref{fig_customized_neural_links}) for class agnostic mask and Single-Shot-Multi-Box-Detector (SSD) \cite{ssd} for box annotation. For both of them, we use comprehensive data-augmentation i.e. random hue, saturation, brightness, and contrast all with selection probability of $0.5$, random rotation between $-10^{\circ}$ to $10^{\circ}$, Gaussian blur $\sigma=3$ and a crop size of $512\times 512$. The training hyper-parameters are set to $learning~rate (\eta) = 0.001$, $learning~rate~policy = step$, $gamma = 0.1$, $momentum = 0.90$ and $weight~decay = 0.0001$.
\par
In general, mAP score is preferred to asses the bounding box prediction accuracy. However, in our case, we are more interested in assessing the overlapping of predicted and ground truth boxes due to presence of only one class (item). Thus, instead of mAP, we report mIoU \cite{voc} score for bounding box prediction and same is also reported for the semantic segmentation. The impact of comprehensive data-augmentation on the tutor can be seen clearly in Table \ref{tab_autoanno}. 
%The Fig.\ref{fig_semi_supervised_labelling_eval} shows automatically generated masks for novel items during an actual competition run.
%
%
\subsection{\textbf{Clutter Synthesis and Quick Child Learning}}
We capture $200$ real cluttered images and manually annotate them to evaluate the performance of quick learning for semantic segmentation. Our focus in this experiment remains on quick learning and therefore we don't involve analysis on large datasets (e.g. \cite{cityscapes}). We evaluate the child network against the state-of-art PSPNet \cite{pspnet} by training both of them for $35$ minutes with a batch size of $5$ on the given computing platform. We freeze the Batch-Normalization \cite{bn} parameters of the back-bone while they are learnt for all the Batch-Normalization layers of the CNN head. We set the training hyper parameters $learning~rate~policy = step$, $gamma = 0.1$, $momentum = 0.90$ and $weight~decay = 0.0001$ and use multinomial softmax loss in which multiple classes compete against each other. The ResNet-$50$ backbone is pretrained for class agnostic segmentation (tutor). To analyze the effect of synthetic clutter and learning rate, we adopt two values of $learning~rates~(\eta) = \{ 0.01, 0.001\}$ and for each of them, the child and PSPNet are trained for varying amounts of clutter.
\par
\begin{figure}[t]
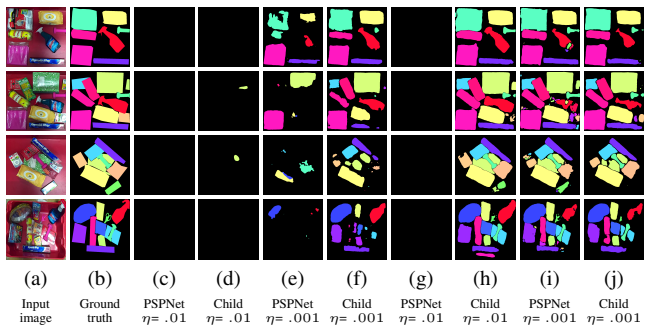

\vspace*{0.3ex}

\centering

\FPeval{\width}{5ex}
\FPeval{\height}{5ex}
\FPeval{\shift}{0.4ex}

\begin{tikzpicture}

\foreach \i in {1,...,4}
\node (node_\i) [rectangle, xshift=0 * (\width +\shift),yshift=-\i * (\height+\shift)]{
\includegraphics[width=\width,height=\height]{images/seg_results_exp/foscam_clutter_\i.png}};

\foreach \i in {1,...,4}
\node (node_\i) [rectangle, xshift=1 * (\width +\shift),yshift=-\i * (\height+\shift)]{
\includegraphics[width=\width,height=\height]{images/seg_results_exp/foscam_clutter_gnd_truth_\i.png}};

\foreach \i in {1,...,4}
\node (node_\i) [rectangle, xshift=2 * (\width +\shift),yshift=-\i * (\height+\shift)]{
\includegraphics[width=\width,height=\height]{images/seg_results_exp/foscam_clutter_mask_01_\i.png}};

\foreach \i in {1,...,4}
\node (node_\i) [rectangle, xshift=3 * (\width +\shift),yshift=-\i * (\height+\shift)]{
\includegraphics[width=\width,height=\height]{images/seg_results_exp/foscam_clutter_mask_FPN_01_\i.png}};

\foreach \i in {1,...,4}
\node (node_\i) [rectangle, xshift=4 * (\width +\shift),yshift=-\i * (\height+\shift)]{
\includegraphics[width=\width,height=\height]{images/seg_results_exp/foscam_clutter_mask_001_\i.png}};

\foreach \i in {1,...,4}
\node (node_\i) [rectangle, xshift=5 * (\width +\shift),yshift=-\i * (\height+\shift)]{
\includegraphics[width=\width,height=\height]{images/seg_results_exp/foscam_clutter_mask_FPN_001_\i.png}};

\foreach \i in {1,...,4}
\node (node_\i) [rectangle, xshift=6 * (\width +\shift),yshift=-\i * (\height+\shift)]{
\includegraphics[width=\width,height=\height]{images/seg_results_exp/foscam_clutter_mask_cluttered_01_\i.png}};

\foreach \i in {1,...,4}
\node (node_\i) [rectangle, xshift=7 * (\width +\shift),yshift=-\i * (\height+\shift)]{
\includegraphics[width=\width,height=\height]{images/seg_results_exp/foscam_clutter_mask_FPN_cluttered_01_\i.png}};

\foreach \i in {1,...,4}
\node (node_\i) [rectangle, xshift=8 * (\width +\shift),yshift=-\i * (\height+\shift)]{
\includegraphics[width=\width,height=\height]{images/seg_results_exp/foscam_clutter_mask_cluttered_001_\i.png}};

\foreach \i in {1,...,4}
\node (node_\i) [rectangle, xshift=9 * (\width +\shift),yshift=-\i * (\height+\shift)]{
\includegraphics[width=\width,height=\height]{images/seg_results_exp/foscam_clutter_mask_FPN_cluttered_001_\i.png}};

\FPeval{\subfloatcaptiony}{0-4.5}

\node (node_1) [rectangle, xshift=0 * (\width +\shift),yshift=\subfloatcaptiony * (\height+\shift), text width= 1*\width-1ex, anchor=north,align=center ]{\footnotesize (a)};
\node (node_2) [rectangle, xshift=1 * (\width +\shift),yshift=\subfloatcaptiony * (\height+\shift), text width= 1*\width-1ex,anchor=north,align=center]{\footnotesize (b)};
\node (node_3) [rectangle, xshift=2 * (\width+\shift),yshift=\subfloatcaptiony * (\height+\shift), text width= 1*\width-1ex, anchor=north,align=center]{\footnotesize (c)};
\node (node_4) [rectangle, xshift=3 * (\width +\shift),yshift=\subfloatcaptiony * (\height+\shift), text width= 1*\width-1ex, anchor=north,align=center]{\footnotesize (d)};
\node (node_5) [rectangle, xshift=4 * (\width +\shift),yshift=\subfloatcaptiony * (\height+\shift), text width= 1*\width-1ex, anchor=north,align=center]{\footnotesize (e)};
\node (node_6) [rectangle, xshift=5 * (\width +\shift),yshift=\subfloatcaptiony * (\height+\shift), text width= 1*\width-1ex, anchor=north,align=center]{\footnotesize (f)};
\node (node_7) [rectangle, xshift=6 * (\width +\shift),yshift=\subfloatcaptiony * (\height+\shift), text width= 1*\width-1ex, anchor=north,align=center]{\footnotesize (g)};
\node (node_8) [rectangle, xshift=7 * (\width +\shift),yshift=\subfloatcaptiony * (\height+\shift), text width= 1*\width-1ex, anchor=north,align=center]{\footnotesize (h)};
\node (node_9) [rectangle, xshift=8 * (\width +\shift),yshift=\subfloatcaptiony * (\height+\shift), text width= 1*\width-1ex, anchor=north,align=center]{\footnotesize (i)};
\node (node_10) [rectangle, xshift=9 * (\width +\shift),yshift=\subfloatcaptiony * (\height+\shift), text width= 1*\width-1ex, anchor=north,align=center]{\footnotesize (j)};

\FPeval{\ycaptionup}{5.5}
\FPeval{\ycaptiondown}{4.3}

\node (node_3_up) [rectangle, below of =node_3,yshift=\ycaptionup ex, text width= 1*\width-1ex, anchor=north,align=center]{\tiny PSPNet};

\node (node_4_up) [rectangle,below of =node_4,yshift=\ycaptionup ex, text width= 1*\width-1ex, anchor=north,align=center]{\tiny Child};

\node (node_5_up) [rectangle,below of =node_5,yshift=\ycaptionup ex, text width= 1*\width-1ex, anchor=north,align=center]{\tiny PSPNet};

\node (node_6_up) [rectangle,below of =node_6,yshift=\ycaptionup ex, text width= 1*\width-1ex, anchor=north,align=center]{\tiny Child};

\node (node_7_up) [rectangle,below of =node_7,yshift=\ycaptionup ex, text width= 1*\width-1ex, anchor=north,align=center]{\tiny PSPNet};

\node (node_8_up) [rectangle,below of =node_8,yshift=\ycaptionup ex, text width= 1*\width-1ex, anchor=north,align=center]{\tiny Child};

\node (node_9_up) [rectangle,below of =node_9, yshift=\ycaptionup ex, text width= 1*\width-1ex, anchor=north,align=center]{\tiny PSPNet};

\node (node_10_up) [rectangle,below of =node_10,yshift=\ycaptionup ex, text width= 1*\width-1ex, anchor=north,align=center]{\tiny Child};

%% eta

\node (node_3_up_eta) [rectangle, below of =node_3,yshift=\ycaptiondown ex, anchor=north,align=center]{\tiny $\eta$= $.01$};

\node (node_4_up_eta) [rectangle, below of =node_4,yshift=\ycaptiondown ex, anchor=north,align=center]{\tiny $\eta$= $.01$};

\node (node_5_up_eta) [rectangle, below of =node_5,yshift=\ycaptiondown ex, anchor=north,align=center]{\tiny $\eta$= $.001$};

\node (node_6_up_eta) [rectangle, below of =node_6,yshift=\ycaptiondown ex, anchor=north,align=center]{\tiny $\eta$= $.001$};

\node (node_7_up_eta) [rectangle, below of =node_7,yshift=\ycaptiondown ex,  anchor=north,align=center]{\tiny $\eta$= $.01$};

\node (node_8_up_eta) [rectangle, below of =node_8,yshift=\ycaptiondown ex,  anchor=north,align=center]{\tiny $\eta$= $.01$};

\node (node_9_up_eta) [rectangle, below of =node_9,yshift=\ycaptiondown ex,  anchor=north,align=center]{\tiny $\eta$= $.001$};

\node (node_10_up_eta) [rectangle, below of =node_10,yshift=\ycaptiondown ex, anchor=north,align=center]{\tiny $\eta$= $.001$};

\node (input) [rectangle,below of =node_1,yshift=\ycaptionup ex,anchor=north,align=center]{\tiny Input};
\node (image) [rectangle,below of =node_1,yshift=\ycaptiondown ex,anchor=north,align=center]{\tiny image};

\node (ground) [rectangle,below of =node_2,yshift=\ycaptionup ex,anchor=north,align=center]{\tiny Ground};
\node (truth) [rectangle,below of =node_2,yshift=\ycaptiondown ex,anchor=north,align=center]{\tiny truth};

\end{tikzpicture}
\caption{(c)-(f) training without synthetic clutter, and (g)-(i) with synthetic clutter.}
\label{fig_seg_results}

\vspace*{-0.3ex}

\end{figure}

Table \ref{tab_segmentation_perfm} shows the mIoU of both the architectures on the mentioned real cluttered images. It can be seen that higher learning rate has adverse effects on PSPNet whereas the child network performs significantly better. With synthetic cluttering, both the networks performs well, however the child network outperforms PSPNet with a visible margin. A grid size of $3\times3$ doesn't adds much onto the accuracy, because all the items in the cluttered images are almost isolated. As the grid size is increased, actual cluttered and occlusion situation appears in the cluttered images, resulting in improved mIoU scores, which can also be verified qualitatively by Fig. \ref{fig_seg_results}. All the segmentation masks are are thresholded at $90\%$ confidence in order to demonstrate, how quick the network can achieve higher confidences. Row-$2$ column-(i) marks the presence of small misclassified patches in the output of PSPNet, whereas these are rarely present in the case of child.
\par
\begin{figure}[t]
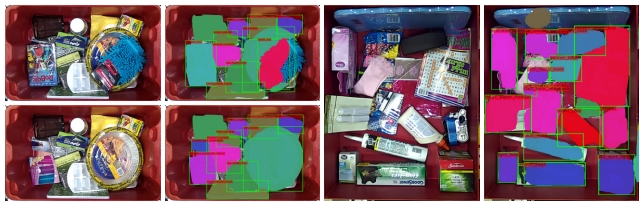

\vspace*{0.3ex}

\centering

\FPeval{\width}{13ex}
\FPeval{\height}{8ex}
\FPeval{\shift}{0.4ex}

\begin{tikzpicture}

\foreach \i in {0,1}
\node (node_\i) [rectangle, xshift=1 * (\width +\shift),yshift=-\i * (\height+\shift)]{
\includegraphics[width=\width,height=\height]{images/official_runs/img_\i.jpg}};

\foreach \i in {0,1}
\node (segnode_\i) [rectangle, xshift=2 * (\width +\shift),yshift=-\i * (\height+\shift)]{
\includegraphics[width=\width,height=\height]{images/official_runs/seg_\i.jpg}};

%\foreach \i in {0,...,0}
%\node (node_\i) [rectangle, xshift=0 * (\width +\shift),yshift=-\i * (\height+\shift)]{
%\includegraphics[width=\width,height=\height]{images/official_runs/img_\i.jpg}};
%
%\foreach \i in {0,...,0}
%\node (segnode_\i) [rectangle, xshift=0 * (\width +\shift),yshift=-1 * (\height+\shift)]{
%\includegraphics[width=\width,height=\height]{images/official_runs/seg_\i.jpg}};

\FPeval{\heightpick}{16.4ex}
\FPeval{\yshiftpick}{0.4ex}

\foreach \i in {7,...,7}
\node (node_\i) [rectangle, xshift= 3* (\width +\shift),yshift=-(\i -7) * (\heightpick+\yshiftpick)-4.25ex]{
\includegraphics[width=\width,height=\heightpick]{images/official_runs/img_\i.jpg}};

\foreach \i in {7,...,7}
\node (segnode_\i) [rectangle, xshift=4 * (\width +\shift),yshift=-(\i-7) * (\heightpick+\yshiftpick)-4.25ex]{
\includegraphics[width=\width,height=\heightpick]{images/official_runs/seg_\i.jpg}};

\FPeval{\subfloatcaptiony}{0-5.5}

%\node (node_a) [rectangle, below of=node_3, yshift=2.5ex, anchor=north,align=center ]{\footnotesize (a)};
%\node (node_b) [rectangle, below of=segnode_3, yshift=2.5ex, anchor=north,align=center ]{\footnotesize (b)};
%\node (node_c) [rectangle, below of=node_8, yshift=-1.20ex, anchor=north,align=center ]{\footnotesize (c)};
%\node (node_d) [rectangle, below of=segnode_8, yshift=-1.20ex, anchor=north,align=center ]{\footnotesize (d)};

\end{tikzpicture}
\caption{Instance detection and segmentation results on the images collected during our ARC'$17$ competition runs}
\label{fig_comptetition_runs}

\vspace*{-0.3ex}

\end{figure}

The quantitative evaluation of instance detection remains same as that of Table \ref{tab_autoanno} because the tutor for box-annotation is employed for instance detection. Fig. \ref{fig_comptetition_runs} shows the qualitative results for instance detection and segmentation. The images were acquired during our actual stow and pick task runs in ARC'$17$.
%
%\begin{table}[h]
%\caption{}
%\centering
%\begin{tabular}{| c | c | c | c | c |}
%
%\hline
%\multirow{3}{*}{Acc(\%)} & \multicolumn{4}{c|}{Synthetic Cluttering} \\ \cline{2-5}   & \multicolumn{2}{c|}{No} & \multicolumn{2}{c|}{Yes}  \\ \cline{2-5} 
%      & $\eta=0.01$ & $\eta=0.001$ & $\eta=0.01$ & $\eta=0.001$ \\ \hline    
%%     Top-1 & 51.295618 & 62.804274  & 51.295618 & 90.852936 & 54.090226 & 82.557583 & 91.747562 & 90.876357  \\ \hline
%%     Top-1 & 51.295618 & 54.090226  & 62.804274 & 82.557583 & 51.295618 & 91.747562 & 90.852936 & 90.876357   \\ \hline
%      %Top-1 & 51.295618 & 51.825345  & 60.789768 & 79.279757 & 51.295618 & 90.821751 & 88.647668 & 88.499037 
%       PSPNet & 51.29 & 59.99 & 62.80 & 84.83  \\ \hline
%       Child & 51.40 & 77.81 & 90.20 & 87.22 \\ \hline
% 
%   \end{tabular}
%%   \hspace{1ex}
%    \label{tab_top1}
%\end{table}
%
\subsection{\textbf{Amazon Robotics Challenge, 2017}}
%\subsubsection{\textbf{No Prior Trained Model}}	
%In contrast to the winner teams \cite{nimbro2017}, \cite{acrvvision2017} who pretrained their network with the the known-set and finetuned for novel items, we assumed each known and novel item to be a novel item. We didn't train any model with known items prior to reaching the competition. Instead, one was trained for both known and novel items on spot. In this process, any human error or gradient explosion and not having any prior model to detect at least the known items, could have led our participation in ARC'$17$ to disqualification. Hence we assured a repeatable, consistence and robust performance by integrating all the components of the vision system with hardware in such a way that it required no manual intervention except the item placing on the rotating platform. In addition, the class labels were automatically inferred from a list according to which the items were being placed for data aquisition process.\\ 
%
\subsubsection{\textbf{Suppressed Misclassification and Open Workspace}}
The on-spot training was done for novel items and the known items, only in the competition set. This strategy allowed the child to penalize the loss function for each item, approximately in a uniform manner. On the other hand, the teams who used their networks pretrained on known items had faced issue of small misclassified patches. It happened due to biasing of their networks towards known items which lead to confusion of novel with known items. In our case, such patches were significantly suppressed and were observed only once during the pick task. Moreover, our system was also robust to ambient lights which allowed us to keep the workspace open and unconstrained (Fig. \ref{fig_full_system}) in contrast to the other teams \cite{nimbro2017}, \cite{acrvvision2017}, \cite{mit2017}.
\subsubsection{\textbf{Inspection Free Self Learning System}}
The improved system starts learning as soon as an item is placed on the platform. It continuously monitors the data acquisition process to examine the number of items processed and generate synthetic clutter only for the items whose images and ground truths are available. In contrast, our vision system in ARC'$17$ coudn't take advantage of the instance detection and online learning. The clutter generation process was, however, runtime, i.e. child never encounters a cluttered image twice. Typically during a training run, the child could learn approximately $\sim67000$ images in $35$ minutes, which is near real time. In addition, all the components of our system were inspection free in contrast to the Team Nimbro and ACRV who manually monitored their data acquisition process and performed correction in case of erroneous ground truth segmentation masks.  %
\subsubsection{\textbf{Statistical Analysis with other teams}}
Our system generated data for itself and exhibited no mis-labeling of an item, just in $45$ minutes. Due to accurate visual perception, we achieved highest grasping accuracies, even more than the winners. Table-\ref{tab_ARC} shows the item grasp success rate for top-5 teams in ARC'$17$. Our team is highlighted in \textcolor{blue}{blue}.
\begin{table}[h]
\centering
\caption{\scriptsize Grasping performance of the Top-$5$ teams in ARC'$17$}
%\begin{tabular}{|>{\tiny}c|>{\tiny}c|>{\tiny}c|>{\tiny}c|} 
\begin{tabular}{c|c|c|c} 

\hline

\multirow{2}{*}{Team} & \multicolumn{3}{c}{Grasp Success Rate} \\ \cline{2-4}

& Stow task & Pick task & Final task \\ \hline

 ACRV & $58.00 ~\%$ & $66.00 ~\%$ & $ 62.50 ~\%  $  \\ 
 NimbRo Picking & $11.11~\%$ &$ 68.40 ~\%$&$ 56.80~\%$ \\
 Nanyang & $38.80 ~\%$ & $\mathbf{100.0} ~\%$ &$ 53.80~\% $ \\
 \textcolor{blue}{IITK-TCS} & \textcolor{blue}{$\mathbf{78.26 ~\%}$} & \textcolor{blue}{$\mathbf{100.0}~\%$} &$ \textcolor{blue}{\mathbf{79.20}~\%} $ \\
  MIT-Princeton & $59.37~ \%$ & $39.00 ~\%$ &$ 64.70~\% $ \\
 \hline

\end{tabular}
\label{tab_ARC} 
\end{table}
%
%\begin{table}[h]
%\centering
%\label{tab:ARC} 
%\begin{tabular}{|c|c|c|c|} 
%\hline
%\multirow{2}{*}{Team} & \multicolumn{2}{c|}{No. of Attempts} & \multirow{2}{*}{Avg Success Rate} \\ \cline{2-3}
%  & Total & Successfull &  \\ \hline
% ACRV &  $17$ & $10$ & $ 62.5 \%$ \\ \hline
% NimbRo Picking & 104 & $49$ & $ 47.11 \%$\\ \hline 
% Nanyang & $85$ & $45$ & $52.9 \%$ \\ \hline
%  \textbf{IITK-TCS} & $57$ & $47$ & $\mathbf{82.45} \%$\\ \hline
%  MIT-Princeton &$141$ & $79$ & $56 \%$\\ \hline
% 
%\end{tabular}
%\end{table}
%
%
\section{Conclusion}
This work introduces real-time technique to autonomously generate high quality box and mask ground truths. It also introduces an occlusion aware clutter synthesis technique which eliminates the need of exhaustive tasks to manually annotate cluttered (multi-class) images. The customized CNN head achieves high accuracies for segmentation in short durations. The learning framework leverages above techniques and have shown outstanding performance in ARC'$17$. After the challenge, the system has been also enabled to detect multiple instances of an item as well as it can learn online, i.e. both labeled data generation and learning happens simultaneously in real time. %In future, we hope to extend this work for object detection and segmentation without treating each object as different category.
\nocite{*}
\bibliographystyle{ieeetr}
\bibliography{main}

\end{document}